\documentclass[11pt,a4paper,english]{article}
\usepackage{a4wide}
\usepackage{graphicx}
\usepackage{amsmath,amssymb,amsthm}
\usepackage{color}

\newtheorem{axiom}{Axiom}

\newtheorem{definition}[axiom]{Definition}

\theoremstyle{definition}

\def\RR{\mathbb{R}}
\def\I{\mathbf{I}}
\def\ZERO{\mathbf{0}}
\def\ONE{\mathbf{1}}
\def\Xcal{{\cal X}}

\def\x{\mathbf{x}}
\def\y{\mathbf{y}}
\def\X{\mathbf{X}}
\def\w{\mathbf{w}}
\def\M{\mathbf{M}}
\def\a{\mathbf{a}}
\def\D{\mathbf{D}}
\def\COV{\mathbf{\Sigma}}
\def\qBar{\bar{q}}
\def\q{\mathbf{q}}
\def\Q{\mathbf{Q}}
\def\alphaV{\boldsymbol{\alpha}}

\newcommand{\normal}[2]{{\cal N}\left(#1,#2\right)}
\newcommand{\Set}[2]{\left\{ \, #1 \, \left| \, #2 \, \right.\right\}}

\newcommand{\Ind}[1]{\mathbb{I}\left\{#1\right\}}  
\newcommand{\norm}[1]{\left\|#1\right\|}
\newcommand{\DOT}[2]{\left\langle#1,#2\right\rangle}

\def\Pr{\mathbb{P}\mbox{r}}
\newcommand{\Prob}[1]{\Pr\left(#1\right)}                        
\newcommand{\Ex}[1]{\mathbb{E}\left[#1\right]}                          
\newcommand{\Ec}[2]{\mathbb{E}\left[\left.\,#1\,\right|\,#2\,\right]}   
\newcommand{\Var}[1]{\mathbb{V}\mbox{ar}\left[#1\right]}                       

\renewcommand{\sec}[1]{Section~\protect\ref{#1}}

\newcommand{\fig}[1]{Figure~\protect\ref{#1}}
\newcommand{\eq}[1]{(\protect\ref{#1})}
\newcommand{\cutout}[1]{}

\newcommand{\substrn}[3]{ #1[#2]}

\newcommand{\z}{{\mathbf{z}}}

\newcommand{\FS}{s}
\newcommand{\andx}{\wedge}

\title{The Feature Importance Ranking Measure\footnote{To appear in the Proceedings of the \emph{European Conference on Machine Learning and Principles and Practice of Knowledge Discovery in Databases (ECML/PKDD)}, 2009.}}
\author{
  Alexander Zien\\Fraunhofer First/Friedrich Miescher Laboratory \\\texttt{alexander.zien@first.fraunhofer.de} \and
  Nicole Kr\"amer\\Berlin Institute of Technology\\\texttt{nkraemer@cs.tu-berlin.de}\and
  S\"oren Sonnenburg\\Friedrich Miescher Laboratory\\\texttt{soeren.sonnenburg@tuebingen.mpg.de}
  \and Gunnar R\"atsch\\Friedrich Miescher Laboratory\\\texttt{gunnar.raetsch@tuebingen.mpg.de}
}
\begin{document}

\maketitle


\begin{abstract}
  Most accurate predictions are typically obtained by learning
  machines with complex feature spaces (as e.g. induced by kernels).
  Unfortunately, such decision rules are hardly accessible to humans
  and cannot easily be used to gain insights about the application
  domain.  Therefore, one often resorts to linear models in
  combination with variable selection, thereby sacrificing some
  predictive power for presumptive interpretability.  Here, we
  introduce the {\em Feature Importance Ranking Measure} (FIRM), which
  by retrospective analysis of arbitrary learning machines allows to
  achieve both excellent predictive performance and superior
  interpretation.  In contrast to standard raw feature weighting, FIRM
  takes the underlying correlation structure of the features into
  account. Thereby, it is able to discover the most relevant features,
  even if their appearance in the training data is entirely prevented
  by noise. The desirable properties of FIRM are investigated
  analytically and illustrated in simulations.
\end{abstract}

\section{Introduction}

A major goal of machine learning --- beyond providing accurate
predictions --- is to gain understanding of the investigated problem.
In particular, for researchers in application areas, it is
frequently of high interest to unveil which features are indicative of
certain predictions.  Existing approaches to the identification of
important features can be categorized according to the restrictions
that they impose on the learning machines.

The most convenient access to features is granted by linear learning
machines.  In this work we consider methods that express their
predictions via a real-valued output function $s:\Xcal\rightarrow\RR$,
where $\Xcal$ is the space of inputs. This includes standard models
for classification, regression, and ranking.  Linearity thus amounts
to
\begin{eqnarray}
  \label{eq:linear}
  s(\x) & = & \w^\top \x + b
  \enspace.
\end{eqnarray}
One popular approach to finding important dimensions of vectorial inputs
($\Xcal=\RR^d$) is {\em feature selection}, by which the
training process is tuned to make sparse use of the available $d$
candidate features.  Examples include $\ell_1$-regularized methods like
Lasso \cite{Tib96} or $\ell_1$-SVMs \cite{BenMan92} and heuristics for
non-convex $\ell_0$-regularized formulations.  They all find feature
weightings $\w$ that have few non-zero components, for example by
eliminating redundant dimensions.  Thus, although the resulting
predictors are economical in the sense of requiring few measurements, it
can not be concluded that the other dimensions are unimportant: a
different (possibly even disjoint) subset of features may yield the same
predictive accuracy.  Being selective among correlated features also
predisposes feature selection methods to be unstable.  Last but not
least, the accuracy of a predictor is often decreased by enforcing
sparsity (see e.g.\ \cite{SonRaeSchSch06}).

In multiple kernel learning (MKL;
e.g.~\cite{LanCriGhaBarJor04,SonRaeSchSch06}) a sparse linear
combination of a small set of kernels \cite{SchSmo02} is optimized
concomitantly to training the kernel machine.  In essence, this lifts
both merits and detriments of the selection of individual features to
the coarser level of feature spaces (as induced by the kernels).  MKL
thus fails to provide a principled solution to assessing the
importance of sets of features, not to speak of individual features.
It is now urban knowledge that $\ell_1$-regularized MKL can even rarely
sustain the accuracy of a plain uniform kernel combination \cite{NIPSMKLWS08}.

Alternatively, the sparsity requirement may be dropped, and the $j$-th
component $w_j$ of the trained weights $\w$ may be taken as the
importance of the $j$-th input dimension.  This has been done, for
instance, in cognitive sciences to understand the differences in human
perception of pictures showing male and female faces
\cite{GraWicBueSch06}; here the resulting weight vector $\w$ is
relatively easy to understand for humans since it can be represented
as an image.

Again, this approach may be partially extended to kernel machines
\cite{SchSmo02}, which do not access the features explicitly.  Instead,
they yield a kernel expansion
\begin{eqnarray}
  \label{eq:kernels}
  s(\x) & = & \sum_{i=1}^n \alpha_i k(\x_i,\x) + b
  \enspace,
\end{eqnarray}
where $(\x_i)_{i=1,\ldots,n}$ are the inputs of the $n$ training
examples.  Thus, the weighting $\alphaV\in\RR^n$ corresponds to the
training examples and cannot be used directly for the interpretation
of features.  It may still be viable to compute explicit weights for
the features $\Phi(\x)$ induced by the kernel via
$k(\x,\x')=\DOT{\Phi(\x)}{\Phi(\x')}$, provided that the kernel is
benign: it must be guaranteed that only a finite and limited number of
features are used by the trained machine, such that the equivalent
linear formulation with
\begin{eqnarray*}
  \w & = & \sum_{i=1}^{n} \alpha_i \Phi(\x_i)
\end{eqnarray*}
can efficiently be deduced and represented.

A generalization of the feature weighting approach that works with
general kernels has been proposed by {\"U}st{\"u}n
et.~al.~\cite{UstMelBuy07}.  The idea is to characterize input
variables by their correlation with the weight vector $\alphaV$.  For
a linear machine as given by (\ref{eq:linear}) this directly results
in the weight vector $\w$; for non-linear functions $s$, it yields a
projection of $\w$, the meaning of which is less clear.

A problem that all above methods share is that the weight that a
feature is assigned by a learning machine is not necessarily an
appropriate measure of its importance.  For example, by multiplying
any dimension of the inputs by a positive scalar and dividing the
associated weight by the same scalar, the conjectured importance of
the corresponding feature can be changed arbitrarily, although the
predictions are not altered at all, i.e.\ the trained learning machine
is unchanged.  An even more practically detrimental shortcoming of the
feature weighting is its failure to take into account correlations
between features; this will be illustrated in a computational
experiment below (Section~\ref{sec:expt}).

Further, all methods discussed so far are restricted to linear scoring
functions or kernel expansions.  There also exists a range of
customized importance measures that are used for building decision
trees and random forests (see e.g.\
\cite{StroBouKneAugZei08,StroBouZeiHot07} for an overview).

In this paper, we reach for an importance measure that is ``universal'':
it shall be applicable to any learning machine, so that we can avoid the
clumsiness of assessing the relevance of features for methods that
produce suboptimal predictions, and it shall work for any feature.  We
further demand that the importance measure be ``objective'', which has
several aspects: it may not arbitrarily choose from correlated features
as feature selection does, and it may not be prone to misguidance by
feature rescaling as the weighting-based methods are.  Finally, the
importance measure shall be ``intelligent'' in that it exploits the
connections between related features (this will become clearer below).

In the next section, we briefly review the state of the art with respect to
these goals and in particular outline a recent proposal, which is,
however, restricted to sequence data.  Section~\ref{sec:firm} exhibits
how we generalize that idea to continuous features and exhibits its
desirable properties.  The next two sections are devoted to unfolding
the math for several scenarios.  Finally, we present a few computational
results illustrating the properties of our approach in the different
settings. The relevant notation is summarized in Table \ref{tab:notation}.

\begin{table}[htb]
\begin{center}
\begin{tabular}{lll}
\hline
symbol & definition & reference\\
\hline
$\mathcal{X}$& input space &\\
$s(x)$& scoring function $\mathcal{X} \rightarrow \mathbb{R}$&\\
$w$& weight vector of a linear scoring function $s$ & equation \eqref{eq:linear}\\
$f$&feature function $\mathcal{X} \rightarrow \mathbb{R}$ &equation \eqref{eq:feature}\\
$q_f(t)$& conditional expected score $\RR \rightarrow \RR$ & definition \ref{def:ces}\\
$Q_f$& feature importance ranking measure (firm) $\in \mathbb{R}$ & definition \ref{def:firm}\\
${\bf{Q}}$ & vector $\in \RR^d$ of firms for $d$ features & subsection \ref{subsec:linear}\\
$\COV,\, \COV_{j\bullet }$&covariance matrix, and its $j$th column&\\
\hline
\end{tabular}
\end{center}
\caption{Notation}
\label{tab:notation}
\end{table}
\subsection{Related Work}  \label{sec:relwork}

A few existing feature importance measures satisfy one or more of the
above criteria.  One popular ``objective'' approach is to assess the
importance of a variable by measuring the decrease of accuracy when
retraining the model based on a random permutation of a variable.
However, it has only a narrow application range, as it is
computationally expensive and confined to input variables.

Another approach is to measure the importance of a feature in terms of
a sensitivity analysis \cite{Fri01}
\begin{eqnarray}
  \label{eq:Ij}
  I_j & = & \Ex{ \left(\frac{\partial s}{\partial x_j}\right)^2 \Var{X_j} }^{1/2}
  \enspace.
\end{eqnarray}
This is both ``universal'' and ``objective''.  However, it clearly
does not take the indirect effects into account: for example, the
change of $X_j$ may imply a change of some $X_k$ (e.g.\ due to
correlation), which may also impact $s$ and thereby augment or
diminish the net effect.

Here we follow the related but more ``intelligent'' idea of
\cite{ZieSonPhiRae08}: to assess the importance of a feature by
estimating its total impact on the score of a trained predictor.
While \cite{ZieSonPhiRae08} proposes this for binary features that
arise in the context of sequence analysis, the purpose of this paper
is to generalize it to real-valued features and to theoretically
investigate some properties of this approach.
It turns out (proof in Section~\ref{sec:normal}) that under normality
assumptions of the input features, FIRM generalizes \eqref{eq:Ij}, as
the latter is a first order approximation of FIRM, and because FIRM
also takes the correlation structure into account.

In contrast to the above mentioned approaches, the proposed {\em
feature importance ranking measure} (FIRM) also takes the dependency
of the input features into account.  Thereby it is even possible to
assess the importance of features that are not observed in the
training data, or of features that are not directly considered by the
learning machine.

\subsection{Positional Oligomer Importance Matrices \cite{ZieSonPhiRae08}}  \label{sec:poims}

In \cite{ZieSonPhiRae08}, a novel feature importance measure called
Positional Oligomer Importance Matrices (POIMs) is proposed for
substring features in string classification.  Given an alphabet
$\Sigma$, for example the DNA nucleotides $\Sigma =
\{{\tt A,C,G,T}\}$, let $\x \in \Sigma^L$ be a sequence of length $L$.
The kernels considered in \cite{ZieSonPhiRae08} induce a feature space
that consists of one binary dimension for each possible substring $\y$
(up to a given maximum length) at each possible position $i$.  The
corresponding weight $w_{\y,i}$ is added to the score if the substring
$\y$ is incident at position $i$ in $\x$.  Thus we have the case of a
kernel expansion that can be unfolded into a linear scoring system:
\begin{eqnarray}
  s(\x) & = & \sum_{\y,i} w_{\y,i} \Ind{ \substrn{\x}{i}{|\y|} = \y }
  \enspace,
\end{eqnarray}
where $\Ind{\cdot}$ is the indicator function.
Now POIMs are defined by
\begin{eqnarray}
  \label{eq:poim}
  Q'(\z,j) & := & \Ec{ \FS(\X) }{ \substrn{\X}{j}{|\z|}=\z } - \Ex{ \FS(\X) }
  \enspace,
\end{eqnarray}
where the expectations are taken with respect to a $D$-th order Markov
distribution.

Intuitively, $Q'$ measures how a feature, here the incidence of
substring $\z$ at position $j$, would change the score $s$ as compared
to the average case (the unconditional expectation).  Although
positional sub-sequence incidences are binary features (they are either
present or not), they posses a very particular correlation structure,
which can dramatically aid in the identification of relevant features.

\section{The Feature Importance Ranking Measure (FIRM)}  \label{sec:firm}

As explained in the introduction, a trained learner is defined by its
output or scoring function $s:\Xcal\rightarrow\RR\,.$ The goal is to
quantify how important any given feature
\begin{eqnarray}
\label{eq:feature}
f:\Xcal&\rightarrow&\RR
\end{eqnarray}
of the input data is to the score. In the case of vectorial inputs
$\Xcal=\RR^d$, examples for features are simple coordinate projections
$f_j(\x)=x_j$, pairs $f_{jk}(\x)=x_{j}x_{k}$ or higher order
interaction features, or step functions
$f_{j,\tau}(\x)=\Ind{x_j>\tau}$ (where $\Ind{\cdot}$ is the indicator
function).

We proceed in two steps. First, we define the expected output of the
score function under the condition that the feature $f$ attains a
certain value.
\begin{definition}[conditional expected score]\label{def:ces}
The  conditional expected score of $s$ for a feature $f$  is the
expected score
$q_f:\RR\rightarrow\RR$ conditional to the feature value $t$ of the
feature $f$:
\begin{eqnarray}
  \label{eq:qft}
  q_f(t) & = &\Ec{s(X)}{f(X)=t}
  \enspace.
\end{eqnarray}
\end{definition}
We remark that this definition corresponds --- up to normalization ---
to the marginal variable importance studied by van der Laan
\cite{Laa06}.  A flat function $q_f$ corresponds to a feature $f$ that
has no or just random effect on the score; a variable function $q_f$
indicates an important feature $f$.

Consequently, the second step of FIRM is to determine the importance
of a feature $f$ as the variability of the corresponding expected score
$q_f:\RR\rightarrow\RR$.
\begin{definition}[feature importance ranking measure]\label{def:firm}
The  feature importance
$Q_f\in\RR$ of the feature $f$ is the standard deviation of the
function $q_f$:
\begin{eqnarray}
  \label{eq:Q}
  Q_f & := & \sqrt{ \Var{q_f(f(X))} }
    = \left( \int_\RR \big( q_f(t) - \qBar_f \big)^2 \Prob{f(X)=t} dt \right)^{\frac{1}{2}}
  \enspace,
\end{eqnarray}
where $\qBar_f := \Ex{q_f(f(X))} = \int_\RR q_f(t) \Prob{f(X)=t} dt$
is the expectation of $q_f$.
\end{definition}

In case of (i) known linear dependence of the score on the feature
under investigation or (ii) an ill-posed estimation problem (\ref{eq:Q})
--- for instance, due to scarce data ---, we suggest to replace the standard
deviation by the more reliably estimated slope of a linear regression.
As we will show later (Section~\ref{sec:bin}), for binary features
identical feature importances are obtained by both ways anyway.

\subsection{Properties of FIRM}

\paragraph{FIRM generalizes POIMs.}
As we will show in Section~\sec{sec:bin}, FIRM indeed contains POIMs
as special case.  POIMs, as defined in (\ref{eq:poim}), are only
meaningful for binary features.  FIRM extends the core idea of POIMs
to continuous features.

\paragraph{FIRM is ``universal''.}
Note that our feature importance ranking measure (FIRM) can be applied
to a very broad family of learning machines.  For instance, it works
in both classification, regression and ranking settings, as long as
the task is modeled via a real-valued output function over the data
points.  Further, it is not constrained to linear functions, as is the
case for $l_1$-based feature selection.  FIRM can be used with any
feature space, be it induced by a kernel or not.  The importance
computation is not even confined to features that are used in the
output function.  For example, one may train a kernel machine with a
polynomial kernel of some degree and afterwards determine the
importance of polynomial features of higher degree.  We illustrate the
ability of FIRM to quantify the importance of unobserved features in
Section~\ref{sec:expt-seq}.

\paragraph{FIRM is robust and ``objective''.}
In order to be sensible, an importance measure is required to be
robust with respect to perturbations of the problem and invariant with respect to
irrelevant transformations.  Many successful methods for
classification and regression are translation-invariant; FIRM will
immediately inherit this property.  Below we show that FIRM is also
invariant to rescaling of the features in some analytically tractable
cases (including all binary features), suggesting that FIRM is
generally well-behaved in this respect.  In Section~\ref{sec:real} we
show that FIRM is even robust with respect to the choice of the learning
method.
FIRM {\em is} sensitive to rescaling of the scoring function $s$.  In
order to compare different learning machines with respect to\ FIRM, $s$ should
be standardized to unit variance; this yields importances $\tilde{Q_f}
=Q_f/\Var{s(X)}^{1/2}$ that are to scale.  Note, however, that the
relative importance, and thus the ranking, of all features for any
single predictor remains fixed.

\paragraph{Computation of FIRM.}
It follows from the definition of FIRM that we need to assess the
distribution of the input features and that we have to compute
conditional distributions of nonlinear transformations (in terms of
the score function $s$). In general, this is infeasible. While in
principle one could try to estimate all quantities empirically, this
leads to an estimation problem due to the limited amount of
data. However, in two scenarios, this becomes feasible.  First, one
can impose additional assumptions. As we show below, for normally
distributed inputs and linear features, FIRM can be approximated
analytically, and we only need the covariance structure of the
inputs. Furthermore, for linear scoring functions \eqref{eq:linear},
we can compute FIRM for (a) normally distributed inputs (b) binary
data with known covariance structure and (c) --- as shown before in
\cite{ZiePhiSon07} --- for sequence data with (higher-order) Markov
distribution.  Second, one can approximate the conditional expected
score $q_f$ by a linear function, and to then estimate the feature
importance $Q_f$ from its slope. As we show in
\sec{sec:bin}, this approximation is exact for binary
data.

\subsection{Approximate FIRM for Normally Distributed Features}  \label{sec:normal}

For general score functions $s$ and arbitrary distributions of the
input, the computation of the conditional expected score
\eqref{eq:qft} and the FIRM score \eqref{eq:Q} is in general
intractable, and the quantities can at best be estimated from the
data. However, under the assumption of normally distributed features,
we can derive an analytical approximation of FIRM in terms of first
order Taylor approximations. More precisely, we use the following
approximation.

{\emph{{\bf Approximation} For a normally random variable $\widetilde X \sim \normal{\widetilde \mu}{\widetilde \COV}$ and a differentiable function $g:\mathbb{R}^d \rightarrow \mathbb{R}^p$, the distribution of $g(X)$ is approximated by its first order Taylor expansion:
\begin{eqnarray*}
g(X) &\sim& \normal{g(\widetilde \mu)}{J \widetilde \COV J^\top}
\end{eqnarray*}
with
\begin{eqnarray*}
J&=& \left.\frac{\partial g}{\partial \x}\right|_{\x=\widetilde \mu}
\end{eqnarray*}
Note that if the function $g$ is linear, the distribution is exact.}}

In the course of this subsection, we consider feature functions
$f_j(\x)=\x_j$ (an extension to linear feature functions $f(\x)=\x^\top
\a$ is straightforward.)

First, recall that for a normally distributed random variable  $X \sim \normal{\ZERO}{\COV}$, the conditional distribution  of  $X|X_j=t$ is again normal, with expectation
\begin{eqnarray*}
\Ec{X}{X_j=t}&=&  \frac{t}{\COV_{jj}}\COV_{j\bullet }=: \widetilde{\bf{\mu}}_j\,.
\end{eqnarray*}
Here $\COV_{j\bullet }$ is the $j$th column of $\COV$.

Now, using the above approximation, the conditional expected score is
\begin{eqnarray*}
q_f(t)&\approx& s\left(\widetilde {\bf{\mu}}_j\right)=s(\left(t/\COV_{jj}\right)  \COV_{j\bullet})
\end{eqnarray*}
To obtain the FIRM score, we apply the approximation again, this time to the function $t \mapsto s((\left(t/\COV_{jj}\right)  \COV_{j\bullet})$. Its first derivative at the expected value $t=0$ equals
\begin{eqnarray*}
J&=& \frac{1}{\COV_{jj}} \COV_{j \bullet} ^\top  \left.\frac{\partial s}{\partial \x}\right|_{\x=\ZERO}
\end{eqnarray*}
This yields
\begin{eqnarray}
\label{eq:qfapprox}
Q_j&\approx& \sqrt{\frac{1}{\COV_{jj} }\left(  \COV_{j \bullet} ^\top  \left.\frac{\partial s}{\partial \x}\right|_{\x=\ZERO}   \right)^2}
\end{eqnarray}
Note the correspondence to \eqref{eq:Ij} in Friedman's paper \cite{Fri01}: If the features are uncorrelated, \eqref{eq:qfapprox} simplifies to
\begin{eqnarray*}
Q_j&\approx& \sqrt{\COV_{jj} \left( \left.\frac{\partial s}{\partial \x_j}\right|_{\x_j=0}     \right)^2}
\end{eqnarray*}
(recall that $0=E[X_j]$).  Hence FIRM adds an additional weighting
that corresponds to the dependence of the input features.  These
weightings are based on the true covariance structure of the
predictors.  In applications, the true covariance matrix is in general
not known.  However, it is possible to estimate it reliably even from
high-dimensional data using mean-squared-error optimal shrinkage \cite{schafer2005sal}.

Note that the above approximation can be used to compute FIRM for the
kernel based score functions \eqref{eq:kernels}. E.g., for Gaussian
kernels
\begin{eqnarray*}
  k_{\gamma}(\x,\x_i)&=& exp\left( -\frac{\|\x - \x_i\|^2}{\gamma^2}  \right)
\end{eqnarray*}
we have
\begin{eqnarray*}
  \left.\frac{\partial k_{\gamma}(\x,\x_i)}{\partial \x}\right|_{\x=0}&=& \frac{2 k(\ZERO,\x_i)}{\gamma^2}  \x_i ^\top = \frac{2 e^{-(\|\x_i\|^2 /\gamma^2)}}{\gamma^2}  \x_i ^\top
\end{eqnarray*}
and hence obtain
\begin{eqnarray*}
  \left.\frac{\partial s}{\partial \x}\right|_{\x=\ZERO} &=& \sum_{i=1} ^N \alpha_i y_i \frac{2 e^{-(\|\x_i\|^2 /\gamma^2)}}{\gamma^2}  \x_i ^\top
  \enspace.
\end{eqnarray*}

\subsection{Exact FIRM for Binary Data}  \label{sec:bin}

Binary features are both analytically simple and, due to their
interpretability and versatility, practically highly relevant.  Many
discrete features can be adequately represented by binary features,
even if they can assume more than two values.  For example, a
categorical feature can be cast into a sparse binary encoding with one
indicator bit for each value; an ordinal feature can be encoded by
bits that indicate whether the value is strictly less than each of its
possibilities. 
Therefore we now try to understand in more depth how FIRM acts on
binary variables.

For a binary feature $f:\Xcal\rightarrow\{a,b\}$ with feature values
$t\in\{a,b\}$, let the distribution be described by
\begin{displaymath}
p_a=\Prob{f(X)=a},\, p_b=1-p_a\,,
\end{displaymath}
 and let the conditional expectations
be $q_a=q_f(a)$ and $q_b=q_f(b)$.  Simple algebra shows that in this
case $\Var{q(f(X))} = p_a p_b (q_a-q_b)^2$.  Thus we obtain the
feature importance
\begin{eqnarray}
  \label{eq:Qbin}
  Q_f & = & ( q_a - q_b ) \sqrt{ p_a p_b }
  \enspace.
\end{eqnarray}
(By dropping the absolute value around $q_a-q_b$ we retain the
directionality of the feature's impact on the score.)  Note that we can interpret firm in terms of the slope of a linear function. If we assume that $a,b \in \mathbb{R}$, the linear regression fit
\begin{eqnarray*}
(w_f,c_f)&=&\text{arg}\min_{w_f,c_f}\int_\RR \left( (w_f t + c_f) - q_f(t) \right)^2
d\Prob{t}
\end{eqnarray*}
the slope is $w_f=\frac{q_a-q_b}{a-b}$.  The
variance of the feature value is $\Var{f(X)} = p_a p_b (a-b)^2$.
\eq{eq:Qbin} is recovered as the increase of the linear regression
function along one standard deviation of feature value.
As desired, the importance is independent of feature translation and
rescaling (provided that the score remains unchanged).  In the
following we can thus (without loss of generality)  constrain that $t\in\{-1,+1\}$.


Let us reconsider POIMS $Q'$, which are defined in equation (\ref{eq:poim}).  We note that $Q'(b) := q_b-\qBar =
p_a(q_b-q_a) = \sqrt{ p_a/p_b } Q(b)$; thus $Q(\z,j)$ can be recovered
as
\begin{eqnarray*}
Q(\z,j)&=& Q'(\z,j)\sqrt{\Prob{\substrn{\X}{j}{|\z|}\neq\z} /
\Prob{\substrn{\X}{j}{|\z|}=\z} }\,.
\end{eqnarray*}
Thus, while POIMs are not strictly a special case of FIRM, they differ
only in a scaling factor which depends on the distribution assumption.
For a uniform Markov model (as empirically is sufficient according to
\cite{ZieSonPhiRae08}), this factor is constant.

\subsection{FIRM for Linear Scoring Functions}
\label{subsec:linear}
To understand the properties of the proposed measure, it is useful to
consider it in the case of linear output functions \eqref{eq:linear}.

\subsubsection{Independently Distributed Binary Data}

First, let us again consider the simplest scenario of uniform binary inputs, $X
\sim unif(\{-1,+1\}^d)$; the inputs are thus pairwise independent.

First we evaluate the importance of the input variables as features,
i.e.~we consider projections $f_j(\x) = x_j$.  In this case, we
immediately find for the conditional expectation $q_j(t)$ of the value
$t$ of the $j$-th variable that $q_j(t) = t w_j + b$.  Plugged into
\eq{eq:Qbin} this yields $Q_j = w_j$, as expected.  When the features
are independent, their impact on the score is completely quantified by
their associated weights; no side effects have to be taken into
account, as no other features are affected.

We can also compute the importances of conjunctions of two variables,
i.e.
\begin{eqnarray*}
f_{j \andx k}(\x) &=& \Ind{ x_j=+1 \andx x_k=+1 }\,.
\end{eqnarray*}
  Here we find
that $q_{j \andx k}(1) = w_j+w_k+b$ and $q_{j \andx k}(0) =
-\frac{1}{3}(w_j+w_k)+b$, with $\Prob{f_{j \andx k}(X)=1} =
\frac{1}{4}$.  This results in the feature importance $Q_{j \andx k}
= (w_j+w_k)/\sqrt{3}$.  This calculation also applies to negated
variables and is easily extended to higher order conjunctions.

Another interesting type of feature derives from the xor-function.  For
features $f_{j \otimes k}(\x) = \Ind{ x_j \neq x_k }$ the conditional
expectations vanish, $q_{j \otimes k}(1) = q_{j \otimes k}(0) = 0$.
Here the FIRM exposes the inability of the linear model to capture
such a dependence.

\subsubsection{Binary Data With Empirical Distribution}

Here we consider the empirical distribution as given by a set
$\Set{\x_i}{i=1,\ldots,n}$ of $n$ data points $\x_i\in\{-1,+1\}^d$:
$\Prob{X} = \frac{1}{n} \sum_{i=1}^{n} \Ind{X=\x_i}$.
For input features $f_j(\x)=x_j$, this leads to
  $q_j(t) = \frac{1}{n_{jt}} \sum_{i:\x_{ij}=t} \w^\top \x_i + b$,
where $n_{jt} := |\Set{i}{\x_{ij}=t}|$ counts the examples showing the
feature value $t$.  With \eq{eq:Qbin} we get
\begin{eqnarray*}
Q_j &=& ( q_j(+1) - q_j(-1) ) \sqrt{ \Prob{X_j=+1} \Prob{X_j=-1} }\\
   &=& \sum_{i=1}^{n} \frac{ \x_{ij} }{ n_{j,\x_{ij}} } \left( \w^\top \x_i
   \right) \sqrt{ \frac{n_{j,+1}n_{j,-1}}{n^2} }
\end{eqnarray*}
It is convenient to express the vector $\Q \in \RR^d$ of all feature
importances in matrix notation.  Let $\X\in\RR^{n \times d}$ be the
data matrix with the data points $\x_i$ as rows.  Then we can write
\begin{eqnarray*}
\Q = \M^\top \X \w&\text{ with } &\M\in\RR^{n \times d}= \ONE_{n \times d} \D_0 + \X \D_1
\end{eqnarray*}
with diagonal matrices $\D_0,\D_1\in\RR^{d \times d}$ defined by\vspace*{-2ex}
\begin{eqnarray}
  (\D_1)_{jj} = \frac{ 1 }{ 2 \sqrt{ n_{j,+1} n_{j,-1} } }
  \enspace,
  & &
  (\D_0)_{jj} = \frac{ n_{j,+1}-n_{j,-1} }{ 2 n \sqrt{ n_{j,+1} n_{j,-1} } }
  \enspace.
\end{eqnarray}
With the empirical covariance matrix $\hat\COV=\frac{1}{n}\X^\top\X$,
we can thus express $\Q$ as
  $\Q = \D_0 \ONE_{d \times n} \X \w + n \D_1 \hat\COV \w$.
Here it becomes apparent how the FIRM, as opposed to the plain $\w$,
takes the correlation structure of the features into account.
Further, for a uniformly distributed feature $j$
(i.e.~$\Prob{X_j=t}=\frac{1}{2}$), the standard scaling is reproduced,
i.e.~$(\D_1)_{jj} = \frac{1}{n}\I$, and the other terms vanish, as
$(\D_0)_{jj} = 0$.

For $\X$ containing each possible feature vector exactly once,
corresponding to the uniform distribution and thus independent
features, $\M^\top\X$ is the identity matrix (the covariance matrix),
recovering the above solution of $\Q=\w$.

\subsubsection{Continuous Data With Normal Distribution}  \label{sec:real}

If we consider normally distributed input features and assume a linear
scoring function \eqref{eq:linear}, the approximations above
(Section~\ref{sec:normal}) are exact.  Hence, the expected conditional
score of an input variable is
\begin{eqnarray}
  q_j(t) & = & \frac{t}{\COV_{jj}} \w^\top \COV_{j\bullet} + b
  \enspace.
\end{eqnarray}
With the diagonal matrix $\D$ of standard deviations of the features,
i.e.~with entries $\D_{jj}=\sqrt{\COV_{jj}}$, this is summarized in
\begin{eqnarray*}
\q &=& b\ONE_d + t\D^{-2}\COV\w \,.
\end{eqnarray*}
Exploiting that the marginal distribution of $X$ with respect to the $j$-th
variable is again a zero-mean normal, $X_j\sim\normal{0}{\COV_{jj}}$,
this yields $\Q = \D^{-1}\COV\w$.
For uncorrelated features, $\D$ is the square root of the diagonal
covariance matrix $\COV$, so that we get $\Q = \D\w$.  Thus rescaling of
the features is reflected by a corresponding rescaling of the
importances --- unlike the plain weights, FIRM cannot be manipulated
this way.

As FIRM weights the scoring vector by the correlation $D^{-1} \Sigma$
between the variables, it is in general more stable and more reliable
than the information obtained by the scoring vector alone.  As an
extreme case, let us consider a two-dimensional variable $(X_1,X_2$)
with almost perfect correlation $\rho=cor(X_1,X_2)\approx1$.  In this
situation, L1-type methods like lasso tend to select randomly only one
of these variables, say $\w=(w_1,0)$, while L2-regularization tends to
give almost equal weights to both variables.  FIRM compensates for the
arbitrariness of lasso by considering the correlation structure of
$X$: in this case $q = (w_1,\rho w_1)$, which is similar to what would
be found for an equal weighting $\w=\frac{1}{2}(w,w)$, namely $q =
(w(1+\rho)/2, w(1+\rho)/2)$.

\paragraph{Linear Regression.}

Here we assume that the scoring function $s$ is the solution of an
unregularized linear regression problem,
$\min_{\w,b} \norm{\X\w-\y}^2$;
thus $\w = \left(\X^\top\X\right)^{-1}\X^\top\y$.

Plugging this into the expression for $\Q$ from above yields
\begin{eqnarray}
  \Q & = & \D^{-1} \COV \left( n\hat\COV \right)^{-1} \X^\top \y
  \enspace.
\end{eqnarray}

For infinite training data, $\hat\COV\longrightarrow\COV$, we thus
obtain $\Q = \frac{1}{n} \D^{-1} \X^\top \y$.
Here it becomes apparent how the normalization makes sense: it renders
the importance independent of a rescaling of the features.  When a
feature is inflated by a factor, so is its standard deviation
$\D_{jj}$, and the effect is cancelled by multiplying them.

\section{Simulation Studies}  \label{sec:expt}

We now illustrate the usefulness of FIRM in a few preliminary
computational experiments on artificial data.

\subsection{Binary Data}

\begin{figure}[htb]
  \vspace{-0.08\textwidth}
  \includegraphics[width=0.33\textwidth]{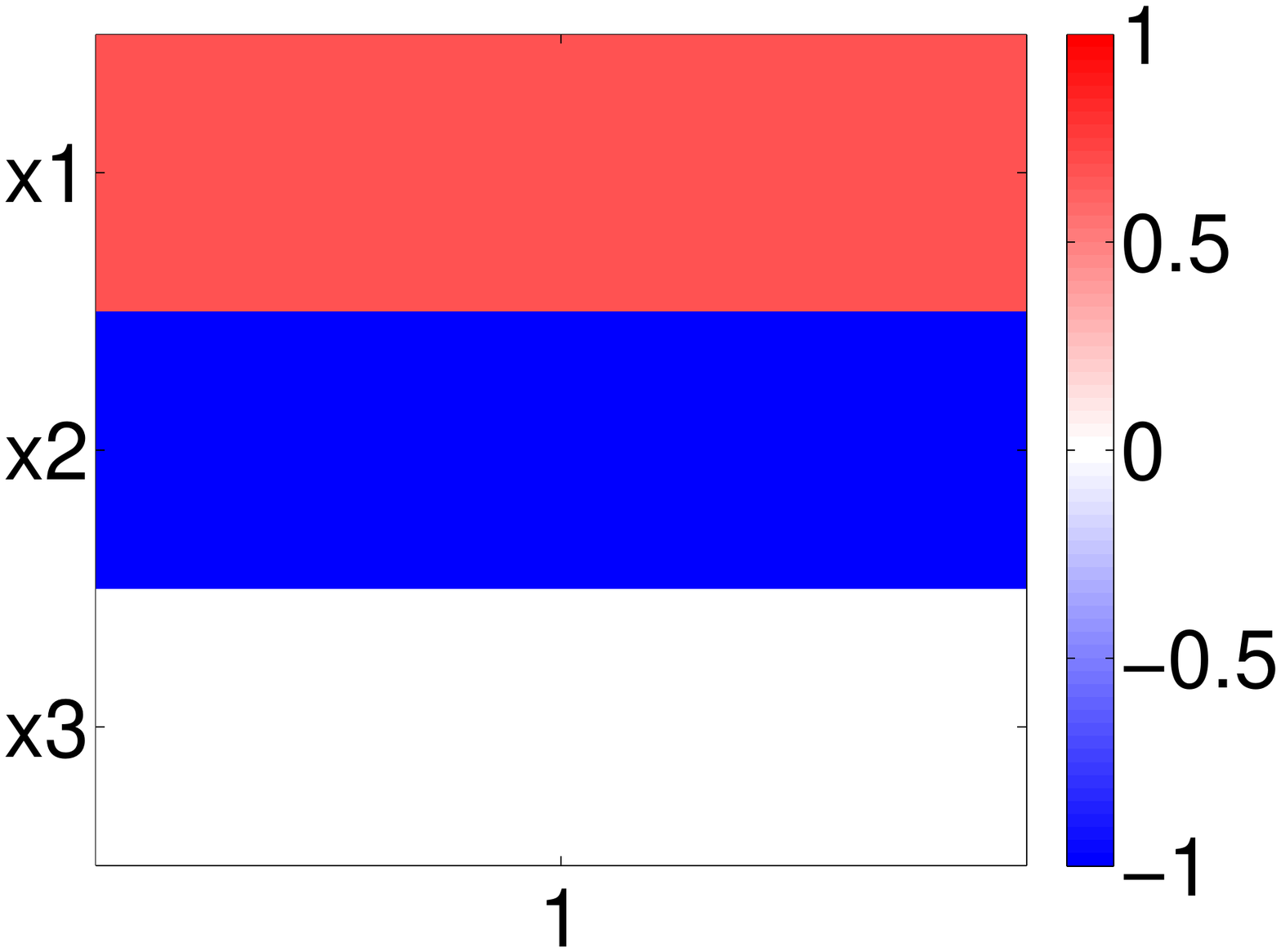}\hspace*{-1mm}
  \includegraphics[width=0.33\textwidth]{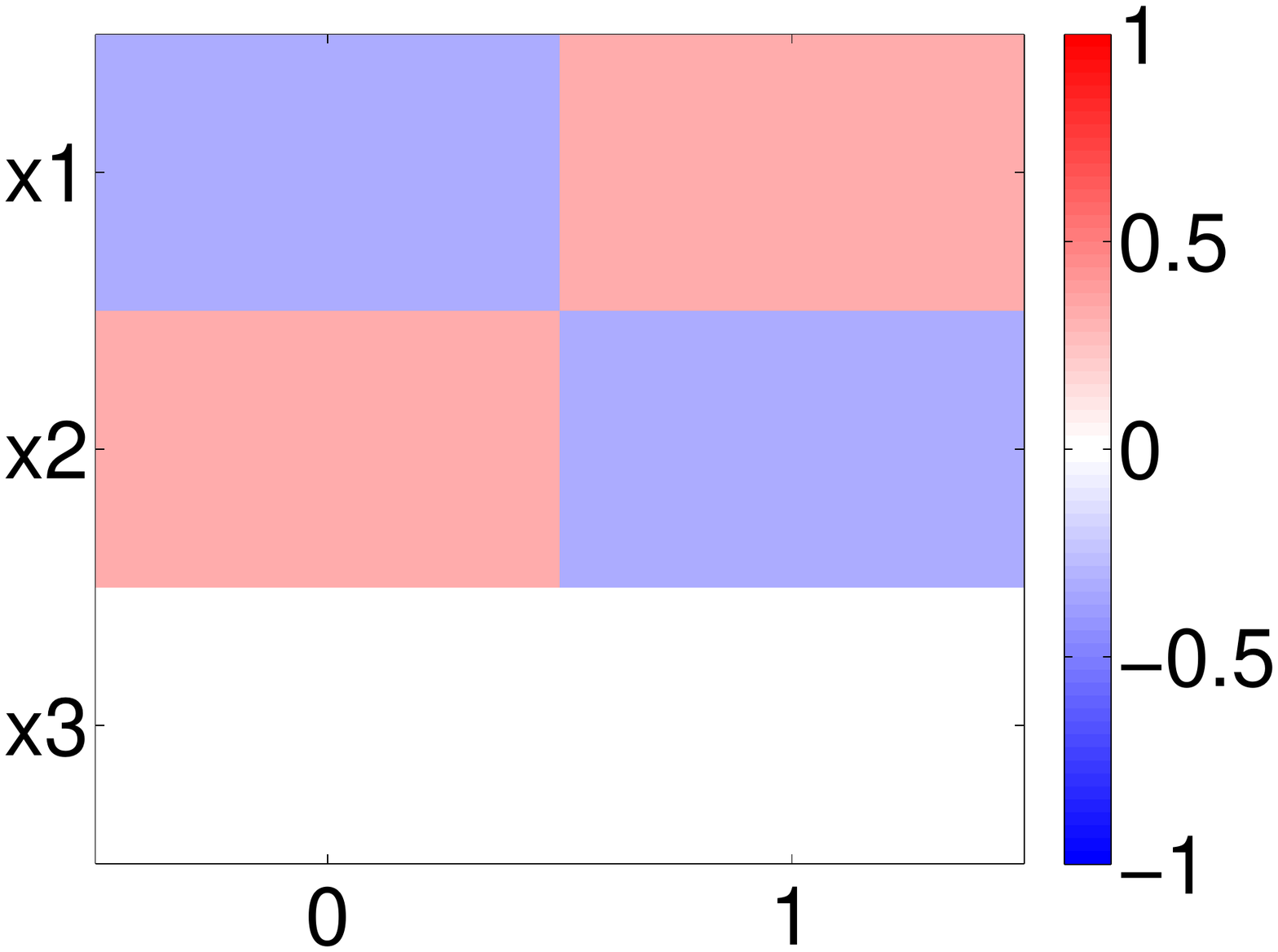}\hspace*{-1mm}
  \includegraphics[width=0.33\textwidth]{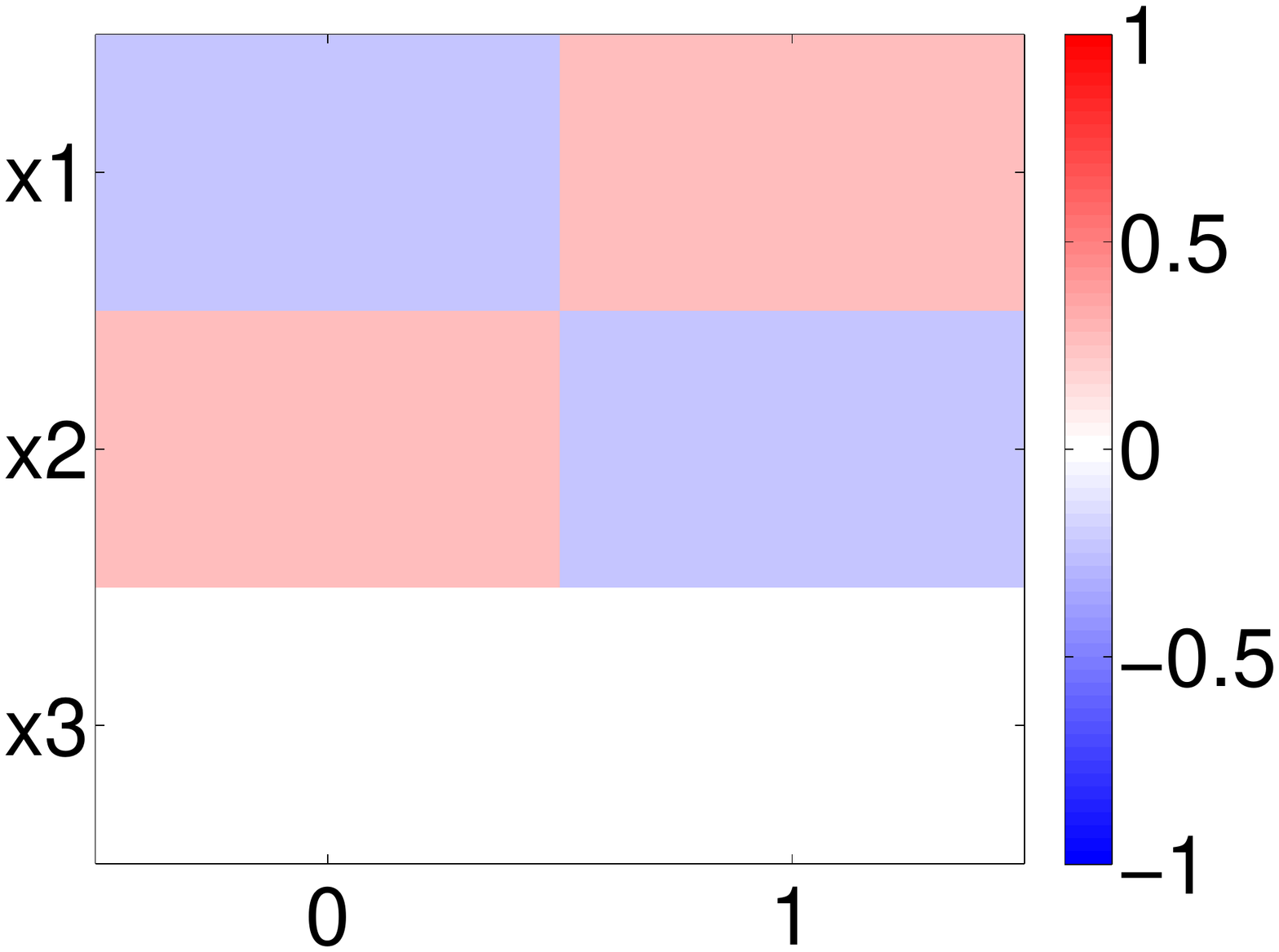}\hspace*{-1mm}
  \vspace{-0.08\textwidth}

  \vspace{-0.08\textwidth}
  \includegraphics[width=0.33\textwidth]{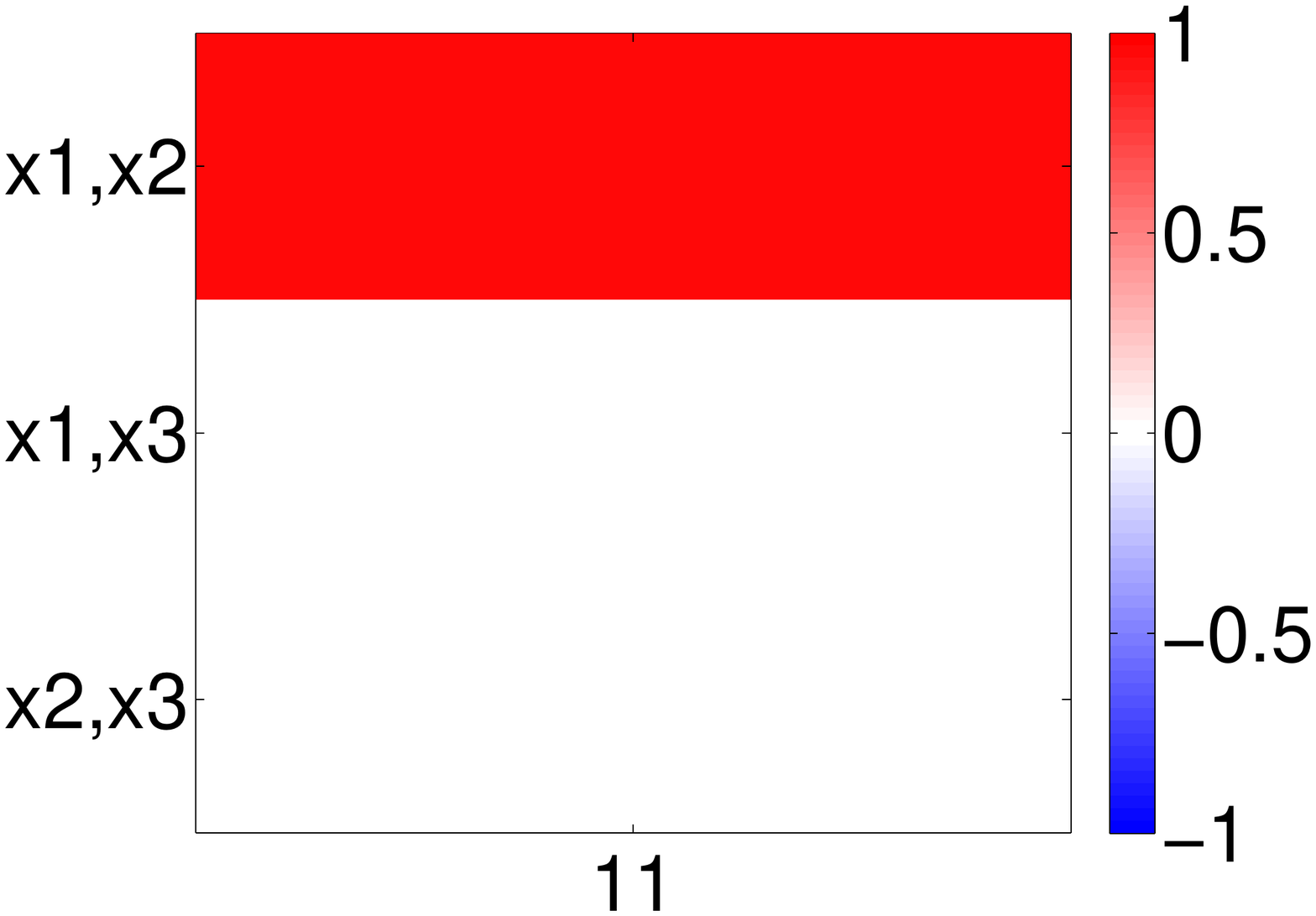}\hspace*{-1mm}
  \includegraphics[width=0.33\textwidth]{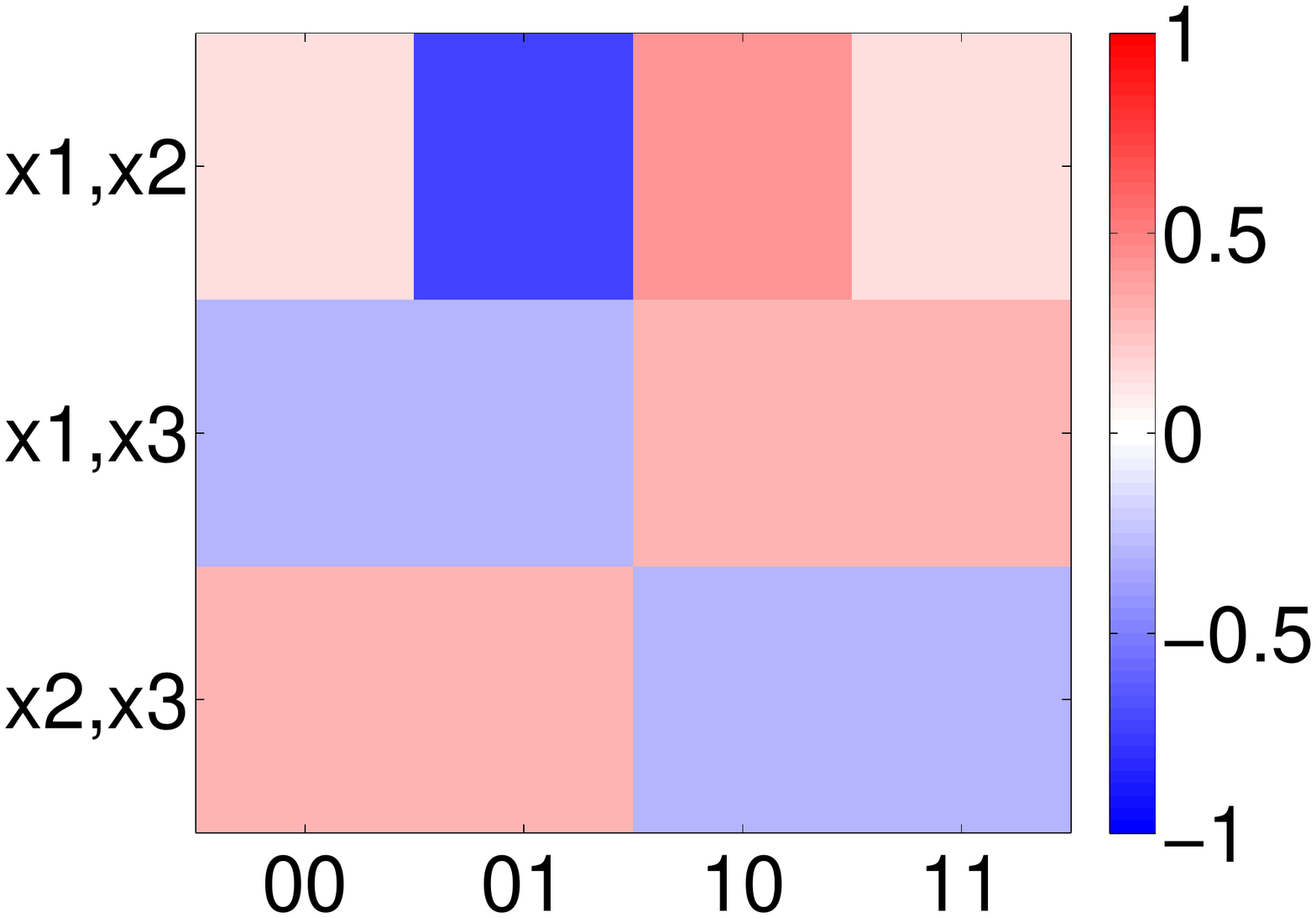}\hspace*{-1mm}
  \includegraphics[width=0.33\textwidth]{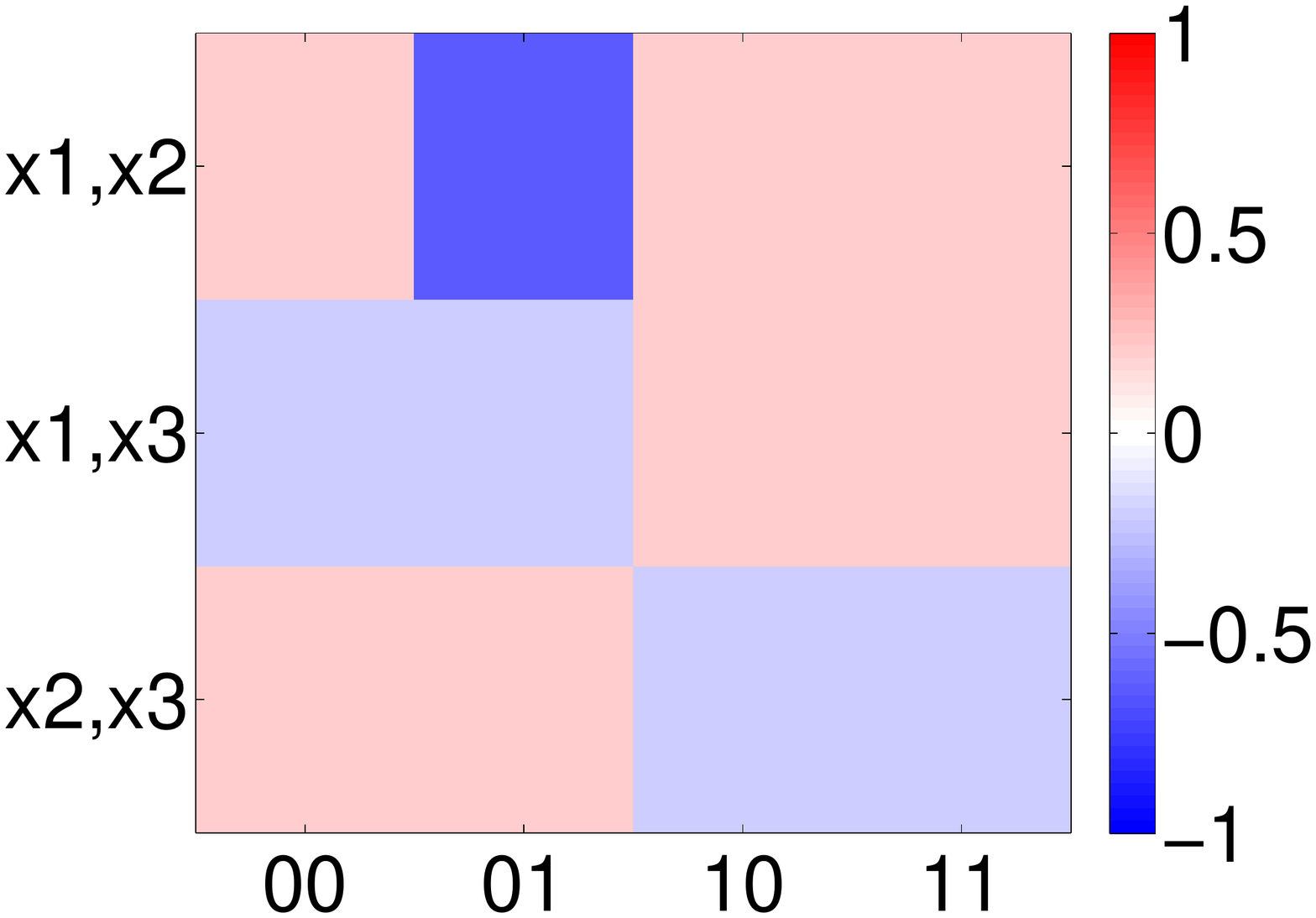}\hspace*{-1mm}

  \vspace{-0.08\textwidth}
  \caption{ \label{fig:bool}
  FIRMs and SVM-$w$ for the Boolean formula $x_1 \vee (\neg x_1 \wedge
  \neg x_2)$.  The figures display heat maps of the scores, blue
  denotes negative label, red positive label, white is neutral.
  The upper row of heat maps shows the scores assigned to a single variable, the
  lower row shows the scores assigned to pairs
  of variables. The first column shows the SVM-$w$ assigning
  a weight to the monomials $x_1, x_2, x_3$ and $x_1x_2, x_1 x_3, x_2
  x_3$ respectively.
  The second column shows FIRMs obtained from the
  trained SVM classifier.  The third column shows FIRMs obtained from the true
  labeling. }\vspace*{-1ex}
\end{figure}

We consider the problem of learning the Boolean formula $x_1 \vee
(\neg x_1 \wedge \neg x_2)$.  An SVM with polynomial kernel of degree
2 is trained on all 8 samples that can be drawn from the Boolean truth
table for the variables $(x_1,x_2,x_3)\in\{0,1\}^3$.  Afterwards, we
compute FIRM both based on the trained SVM ($\w$) and based on the
true labelings ($y$).  The results are displayed in Figure
\ref{fig:bool}.

Note that the raw SVM $\w$ can assign non-zero weights only to
feature space dimensions (here, input variables and their pairwise
conjunctions, corresponding to the quadratic kernel); all other
features, here for example pairwise disjunctions, are implicitly
assigned zero.  The SVM assigns the biggest weight to $x_2$, followed
by $x_1 \wedge x_2$.  In contrast, for the SVM-based FIRM the most
important features are $x_1 \wedge \neg x_2$ followed by $\neg
x_{1/2}$, which more closely resembles the truth.  Note that, due to
the low degree of the polynomial kernel, the SVM not capable of
learning the function ``by heart''; in other words, we have an
underfitting situation.  In fact, we have $s(\x)= 1.\bar{6}$ for
$(x_1,x_2)=(0,1)$.

The difference in $y-$FIRM and SVM-FIRM underlines that --- as intended
--- FIRM helps to understand the learner, rather than the problem.
Nevertheless a quite good approximation to the truth is found as
displayed by FIRM on the true labels, for which all seven 2-tuples
that lead to true output are found (black blocks) and only $\neg x_1
\wedge x_2$ leads to a false value (stronger score).  Values where
$\neg x_1$ and $x_2$ are combined with $x_3$ lead to a slightly
negative value.


\subsection{Gaussian Data}

\begin{figure}[h!]
    \vspace*{-10ex}
    \begin{center}
        \scalebox{1.1}{\hspace*{-5mm}\includegraphics[width=0.29\textwidth]{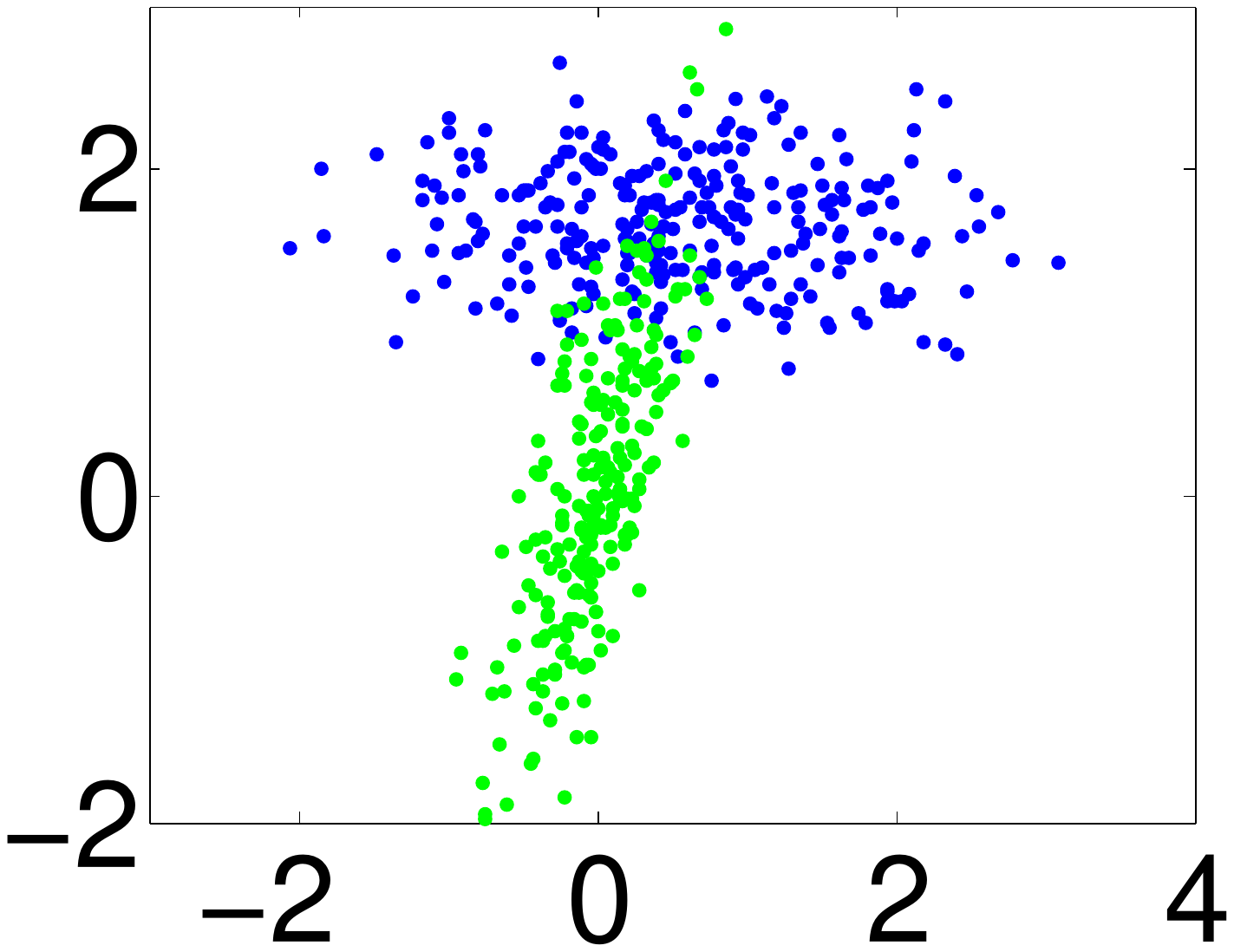}\hspace*{-10mm}
        \includegraphics[width=0.29\textwidth]{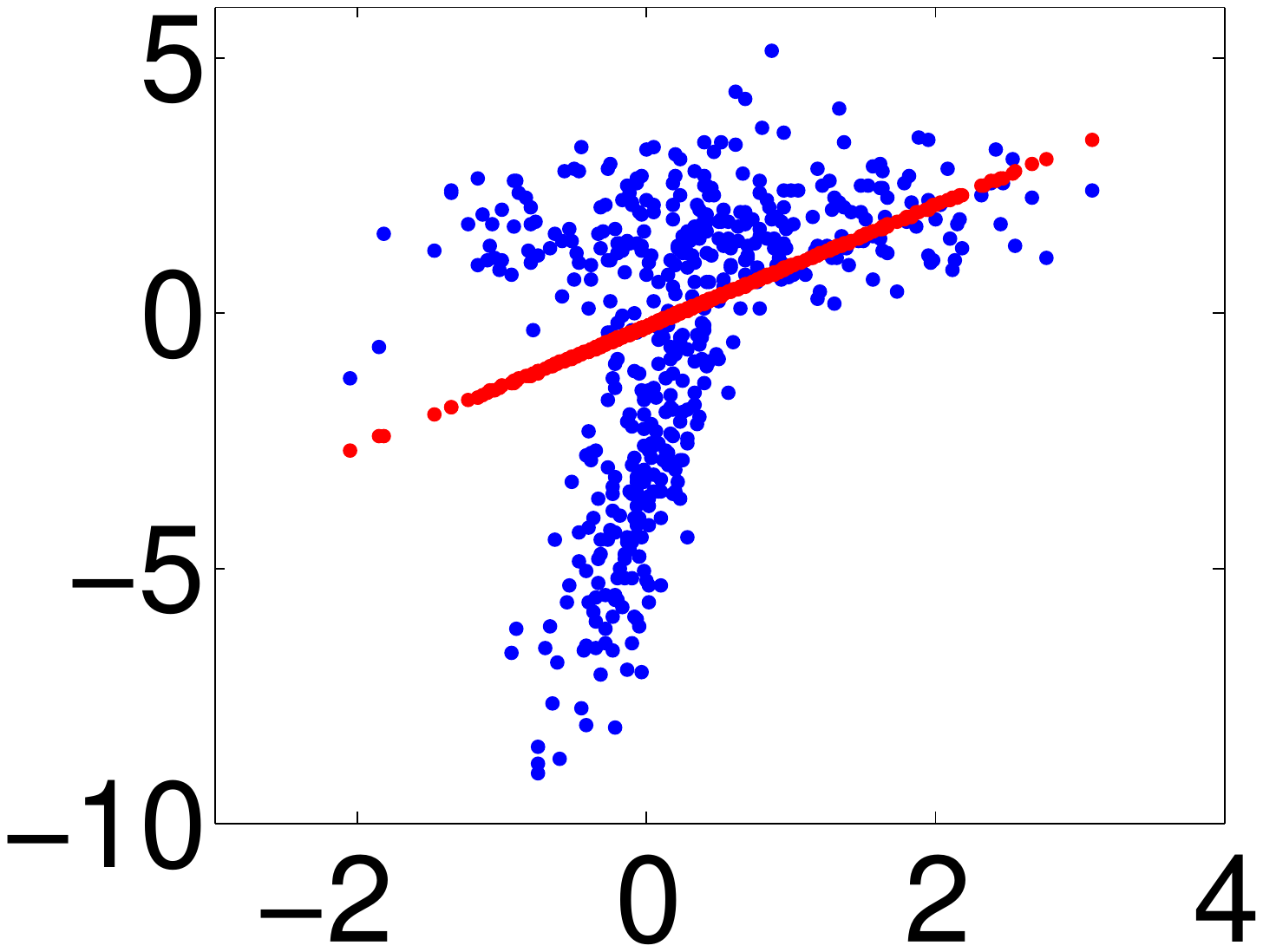}\hspace*{-10mm}
     \includegraphics[width=0.29\textwidth]{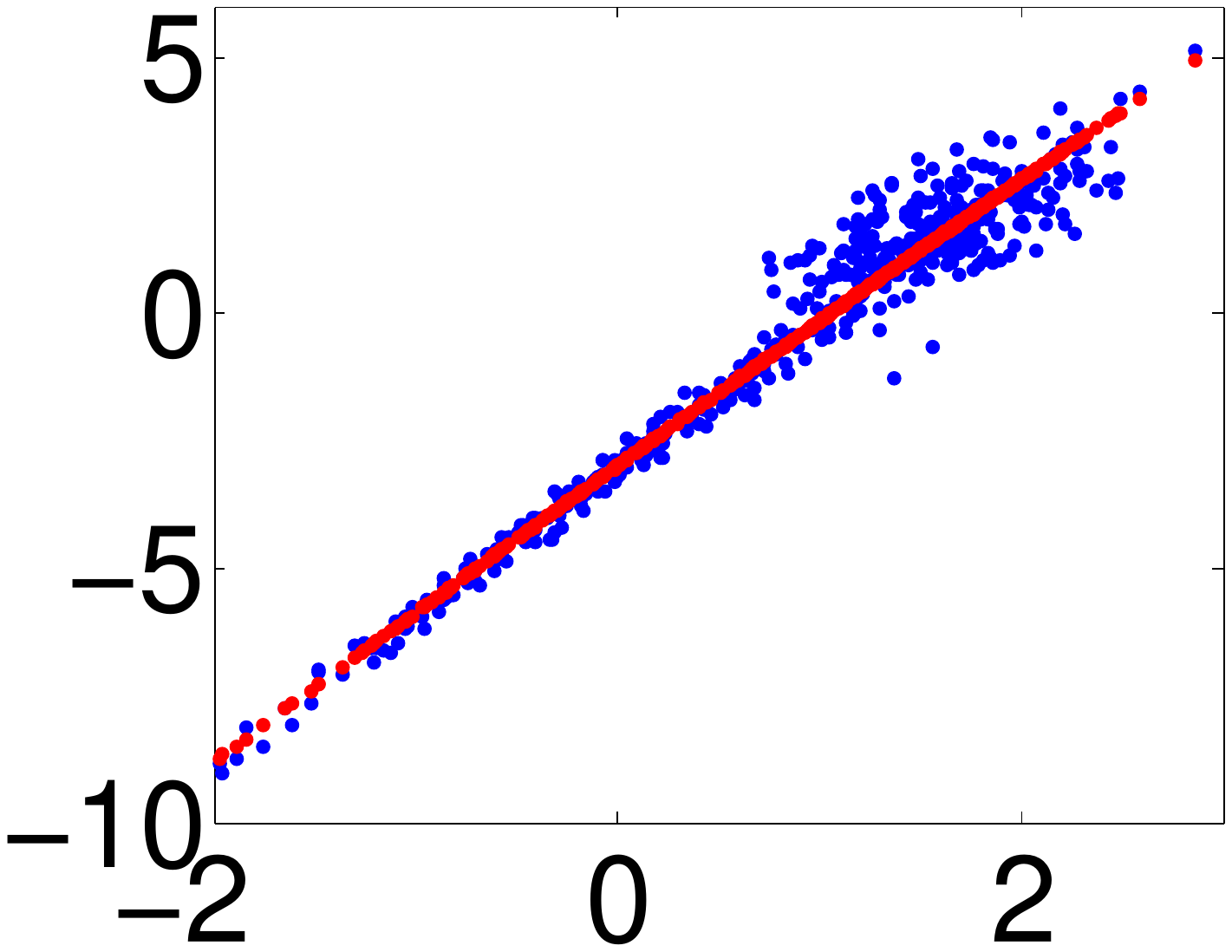}\hspace*{-10mm}
     \includegraphics[width=0.29\textwidth]{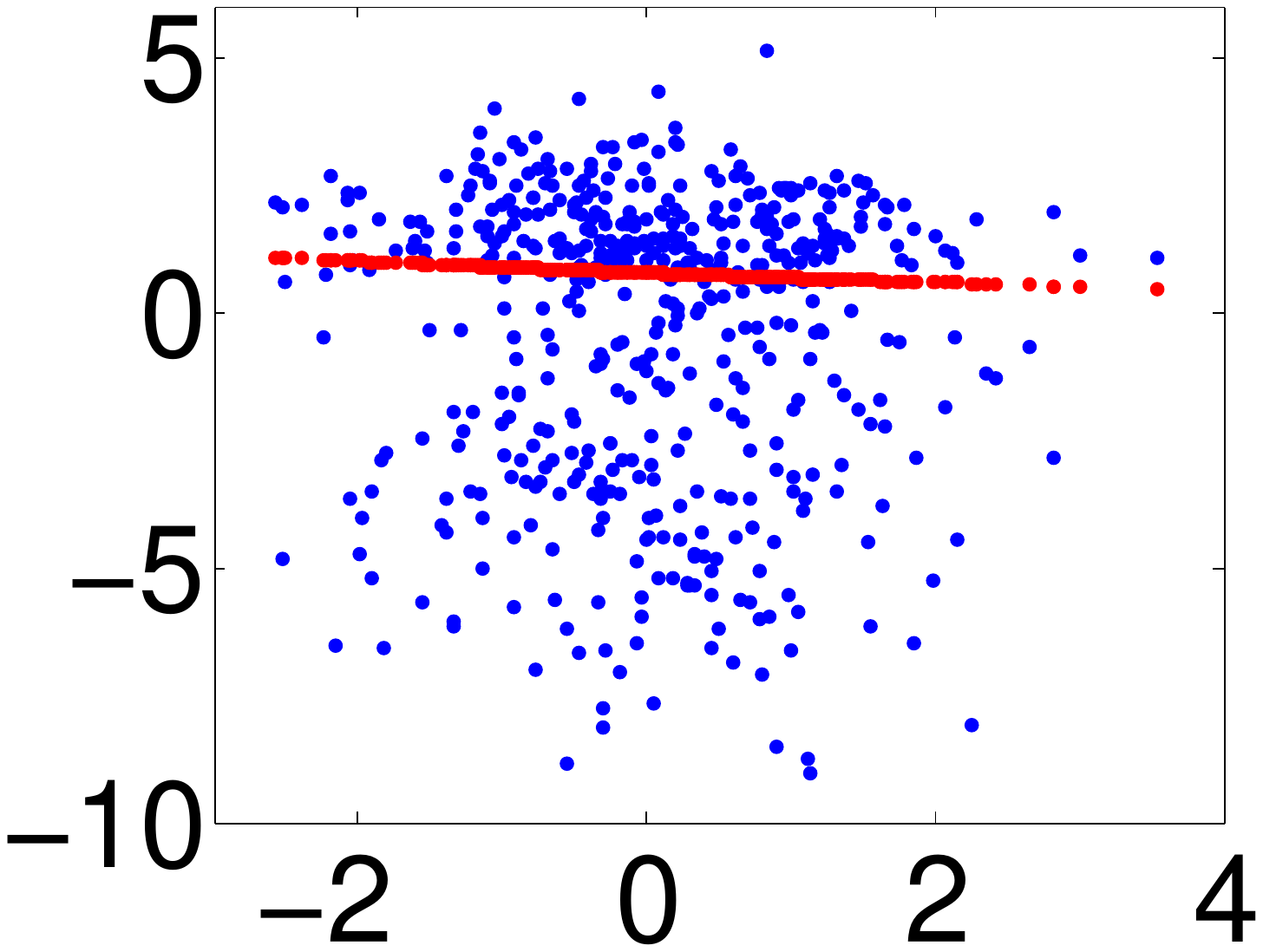}\hspace*{-5mm}}
     \vspace*{-11ex}
    \caption{Binary classification performed on continuous data that consists of two
      3d Gaussians constituting the two classes (with $x_3$ being pure
      noise). From left to right a) Of the raw data set $x_1$, $x_2$
      are displayed. b) Score of the linear discrimination function
      $s(\x_i)$ (blue) and conditional expected score $q_1( (\x_i)_1 )$
      (red) for the first dimension of $\x$. c) $s(\x_i)$ and $q_2(
      (\x_i)_2)$ for varying $x_2$. As the variance of $q$ is highest
      here, this is the discriminating dimension (closely resembling the
      truth). d) $s(\x_i)$ and $q_3( (\x_i)_3)$ for varying $x_3$. Note
      that $x_3$ is the noise dimension and does not contain
      discriminating information (as can be seen from the small slope of
      $q_3$)\label{fig:regression}}
    \end{center}
    \vspace*{-3ex}
\end{figure}

Here, we analyze a toy example to illustrate FIRM for real valued
data. We consider the case of binary classification in three real-valued
dimensions. The first two dimensions carry the discriminative
information (cf.\ Figure~\ref{fig:regression}a), while the third only
contains random noise. The second dimension contains most discriminative
information and we can use FIRM to recover this fact. To do so, we train
a linear SVM classifier to obtain a classification function $s(\x)$. Now
we use the linear regression approach to model the conditional expected
scores $q_i$ (see\ Figure~\ref{fig:regression}b-d for the three
dimensions). We observe that dimension two indeed shows the strongest
slope indicating the strongest discriminative power, while the third
(noise) dimension is identified as uninformative.

\subsection{Sequence Data}  \label{sec:expt-seq}

As shown above (Section~\ref{sec:poims}), for sequence data FIRM is
essentially identical to the previously published technique POIMs
\cite{ZieSonPhiRae08}.  To illustrate its power for sequence
classification, we use a toy data set from \cite{SonRaeSch05}: random
DNA sequences are generated, and for the positive class the
sub-sequence {\tt GATTACA} is planted at a random position centered
around 35 (rounded normal distribution with SD=7).  As biological
motifs are typically not perfectly conserved, the planted consensus
sequences are also mutated: for each planted motif, a single position
is randomly chosen, and the incident letter replaced by a random
letter (allowing for no change for $\sim25\%$ of cases).  An SVM with
WDS kernel \cite{RaeSonSch05} is trained on 2500 positive and as many
negative examples.

\begin{figure}[htpb!]
\begin{center}
  \vspace*{-0.15\columnwidth}
  \includegraphics[width=0.49\columnwidth]{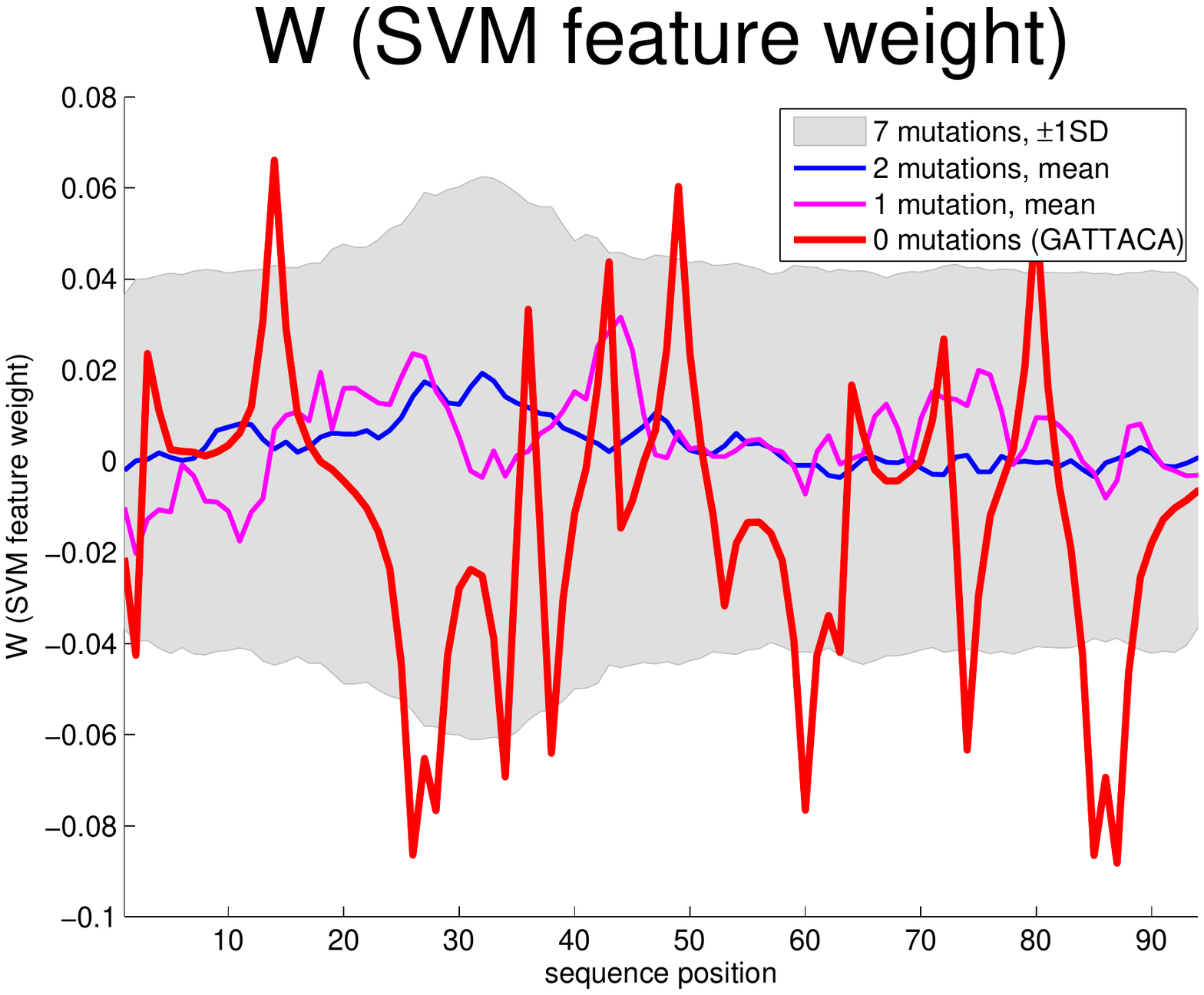}
  \includegraphics[width=0.49\columnwidth]{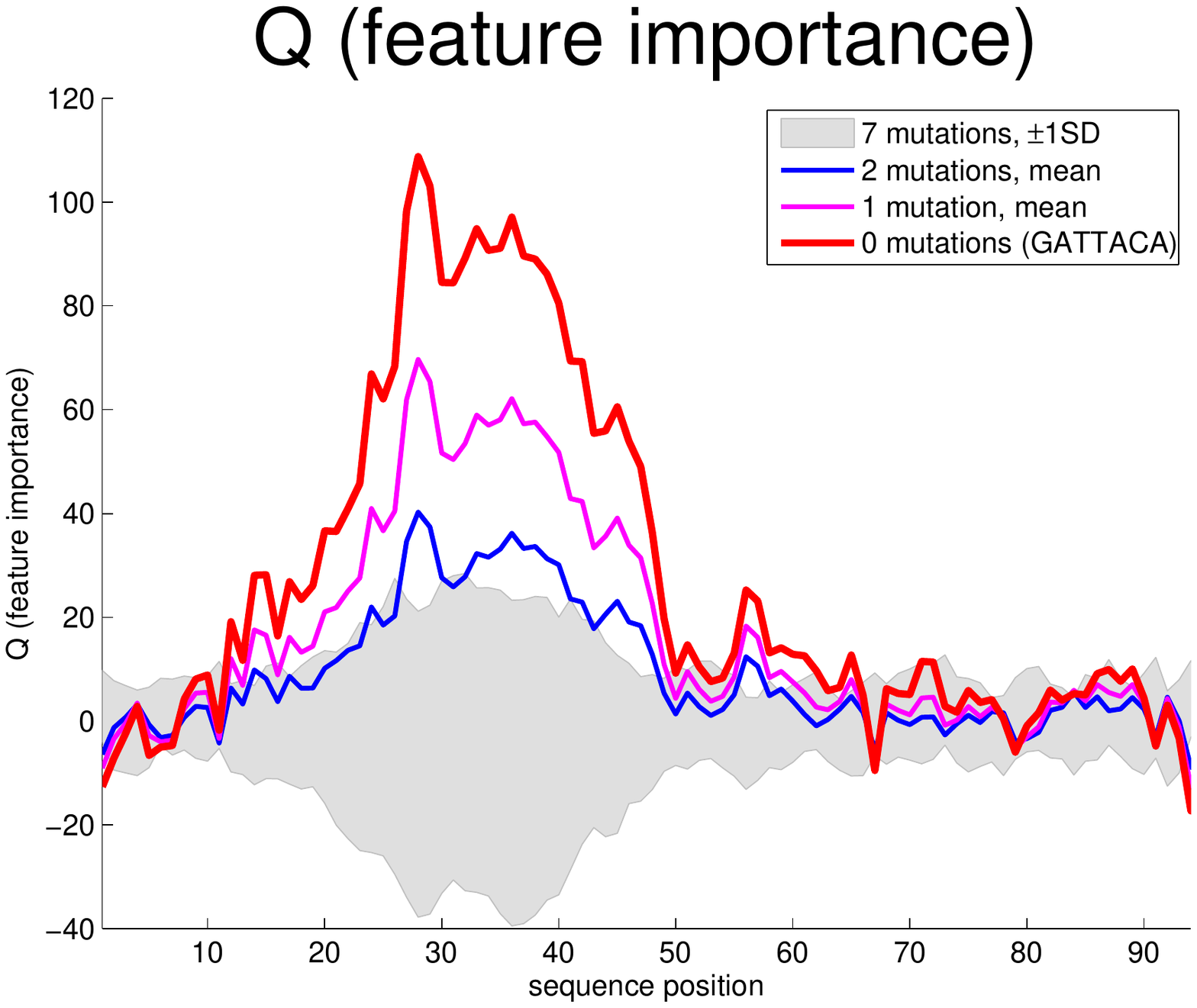}
  \vspace*{-0.15\columnwidth}
  \\
  \caption{ \small \label{fig:pseudo}
    Feature importance analyses based on (left) the SVM feature
    weighting $\w$ and (right) FIRM.  The shaded area shows the $\pm
    1$ SD range of the importance of completely irrelevant features
    (length 7 sequences that disagree to {\tt GATTACA} at every
    position).  The red lines indicate the positional importances of
    the exact motif {\tt GATTACA}; the magenta and blue lines
    represent average importances of all length 7 sequences with edit
    distances 1 and 2, respectively, to {\tt GATTACA}.  While the
    feature weighting approach cannot distinguish the decisive motiv
    from random sequences, FIRM identifies it confidently.
  }
\end{center}
\end{figure}

Two analyses of feature importance are presented in \fig{fig:pseudo}:
one based on the feature weights $\w$ (left), the other on the feature
importance $Q$ (right).  It is apparent that FIRM identifies the {\tt
GATTACA} feature as being most important at positions between 20 and
50, and it even attests significant importance to the strings with
edit distance 1.  The feature weighting $\w$, on the other hand, fails
completely: sequences with one or two mutations receive random
importance, and even the importance of the consensus {\tt GATTACA}
itself shows erratic behavior.

The reason is that the appearance of the exact consensus sequence is
not a reliable feature, as is mostly occurs mutated.  More useful
features are substrings of the consensus, as they are less likely to
be hit by a mutation.  Consequently there is a large number of such
features that are given high weight be the SVM.  By taking into
account the correlation of such short substrings with longer ones, in
particular with {\tt GATTACA}, FIRM can recover the ``ideal'' feature
which yields the highest SVM score.  Note that this ``intelligent''
behavior arises automatically; no more domain knowledge than the
Markov distribution (and it is only $0$-th order uniform!) is
required.  The practical value of POIMs for real world biological
problems has been demonstrated in \cite{ZieSonPhiRae08}.

\section{Summary and Conclusions}

We propose a new measure that quantifies the relevance of features. We take up the idea underlying a recent sequence analysis method
(called POIMs, \cite{ZieSonPhiRae08}) --- to assess the importance of
substrings by their impact on the expected score --- and generalize it
to arbitrary continuous features.  The resulting {\em feature
importance ranking measure} FIRM has invariance properties that are
highly desirable for a feature ranking measure.
First, it is ``objective'': it is invariant with respect to
translation, and reasonably invariant with respect to rescaling of the
features.
Second, to our knowledge FIRM is the first feature ranking measure
that is totally ``universal'', i.e.\ which allows for evaluating {\em
any} feature, irrespective of the features used in the primary
learning machine.  It also imposes no restrictions on the learning
method.
Most importantly, FIRM is ``intelligent'': it can identify features
that are not explicitly represented in the learning machine, due to
the correlation structure of the feature space.  This allows, for
instance, to identify sequence motifs that are longer than the
considered substrings, or that are not even present in a single
training example.

By definition, FIRM depends on the distribution of the input features, which is in general not available. We showed that under various scenarios (e.g. binary features, normally distributed features), we can obtain approximations of FIRM that can be efficiently computed from data. In real-world scenarios, the underlying assumptions might not always be fulfilled. Nevertheless, e.g. with respect to the normal distribution, we can still interpret the derived formulas as an estimation based on first and second order statistics only.

While the quality of the computed importances does depend on the
accuracy of the trained learning machine, FIRM can be used with any
learning framework.  It can even be used without a prior learning
step, on the raw training data.  Usually, feeding training labels as
scores into FIRM will yield similar results as using a learned
function; this is natural, as both are supposed to be highly
correlated.

However, the proposed indirect procedure may improve the results due
to three effects: first, it may smooth away label errors; second, it
extends the set of labeled data from the sample to the entire space;
and third, it allows to explicitly control and utilize distributional
information, which may not be as pronounced in the training sample.  A
deeper understanding of such effects, and possibly their exploitation
in other contexts, seems to be a rewarding field of future research.

Based on the unique combination of desirable properties of FIRM, and
the empirical success of its special case for sequences, POIMs
\cite{ZieSonPhiRae08}, we anticipate FIRM to be a valuable tool for
gaining insights where alternative techniques struggle.

\subsubsection*{Acknowledgements}
This work was supported in part by the FP7-ICT Programme of the European Community under the PASCAL2 Network of Excellence (ICT-216886), by the Learning and Inference Platform of the Max Planck and Fraunhofer Societies, and by the BMBF grant FKZ 01-IS07007A (ReMind). We thank Petra Philips for early phase discussion.

\end{document}